\documentclass[runningheads]{llncs}

\usepackage[review,year=2026,ID=*****]{eccv}
\makeatletter
\@ifundefined{maketitleold}{}{%
  \let\maketitle\maketitleold
}
\makeatother

\usepackage{eccvabbrv}

\usepackage{graphicx}
\usepackage{booktabs}
\usepackage{amsmath,amssymb}
\usepackage{mathtools}
\usepackage{microtype}
\usepackage{subcaption}
\usepackage{enumitem}
\usepackage{multirow}
\usepackage{float}
\usepackage{wrapfig}
\usepackage{placeins}
\usepackage{xurl}
\usepackage{tabularx}
\usepackage{array}
\usepackage[accsupp]{axessibility}
\usepackage{hyperref}
\usepackage[capitalize,noabbrev]{cleveref}
\usepackage[textsize=tiny]{todonotes}

\begin{document}

\makeatletter
\@ifundefined{nolinenumbers}{}{\nolinenumbers}
\makeatother

\title{FuzzingRL: Reinforcement Fuzz-Testing for Revealing VLM Failures}
\titlerunning{FuzzingRL}

\author{
Jiajun Xu\inst{1} \and
Jiageng Mao\inst{1}\textsuperscript{\textdagger} \and
Ang Qi\inst{1} \and
Weiduo Yuan\inst{1} \and
Alexander Romanus\inst{1} \and
Helen Xia\inst{1} \and
Vitor Campagnolo Guizilini\inst{2} \and
Yue Wang\inst{1}\textsuperscript{\textdagger}
}

\institute{
University of Southern California \and
Toyota Research Institute
}

\maketitle

\vspace{-1.8em}
\begin{center}
\small
\textsuperscript{\textdagger}Equal advising.
\end{center}
\vspace{-0.9em}

\begin{abstract}
    Vision-Language Models (VLMs) are prone to errors, and identifying where these errors occur is critical for ensuring the reliability and safety of AI systems. In this paper, we propose an approach that automatically generates questions designed to deliberately induce incorrect responses from VLMs, thereby revealing their vulnerabilities. The core of this approach lies in fuzz-testing and reinforcement fine-tuning: we transform a single input query into a large set of diverse variants through vision and language fuzzing. Based on the fuzzing outcomes, the question generator is further instructed by adversarial reinforcement fine-tuning to produce increasingly challenging queries that trigger model failures. With this approach, we can consistently drive down a target VLM’s answer accuracy—for example, the accuracy of Qwen2.5-VL-32B on our generated questions drop from 86.58\% to 65.53\% in four RL iterations. Moreover, a fuzzing policy trained against a single target VLM transfers to multiple other VLMs, producing challenging queries that degrade their performance as well.
\end{abstract}

\section{Introduction}

Recently, Vision–Language Models (VLMs) have been advancing rapidly and are now widely adopted as the visual backbone in domains such as Vision–Language–Action (VLA) systems, world models, and AI agents. However, issues such as uneven token weight distribution in attention mechanisms, textual bias in the LLM component, and misalignment between the vision encoder and the language model have led to various forms of hallucination in VLMs. As VLMs become the core component of multimodal systems and autonomous agents, their errors can directly cause decision failures, bias propagation, or even safety risks. Therefore, ensuring the robustness and controllability of VLMs is crucial for the development of trustworthy multimodal intelligence.

Therefore, discovering failures of VLMs becomes increasingly important. Prior work has primarily designed static benchmarks that assess model capability with specific items, including broad coverage suites \cite{yue2024mmmu,yue2024mmmu_pro} , multi-dimensional objective evaluations \cite{liu2024mmbench,zhang2024mme,li2024seed}, assessments closer to open domain QA and complex scenes \cite{yu2023mm-vet,liu2023llava-next}, as well as specialized datasets \cite{lu2023mathvista,saikh2022scienceqa} and defect-diagnostic lines \cite{zhao2022vlchecklist,mai2025avabench,zhang2025negvqa,li2024naturalbench}. While these efforts substantially improve coverage and diagnosability, mainstream evaluations often require humans to first identify specific shortcomings of the VLM, and then construct a benchmark targeting those weaknesses. They are largely static and rely heavily on humans to manually reveal VLM failures, making it difficult to adaptively focus on the truly high-failure regions within the vast vision–language combinatorial space.

This raises a fundamental question: \textit{can we design a framework that autonomously discovers failures in VLMs?} To explore this, we decompose the problem into two key challenges. First, we must generate inputs to the VLM that are sufficiently diverse to ensure broad coverage of the vast vision–language space. Second, given the large set of generated question answer pairs, we need a mechanism to adaptively guide the input generation process toward the model’s most vulnerable regions, and continuously refine it to produce increasingly challenging queries that can reliably expose VLM failures.

To this end, we propose FuzzingRL, a framework that aims to automatically expose and amplify the weaknesses of vision-language models. The framework consists of two synergistic components: vision-language fuzzing for systematic input diversification and adversarial reinforcement finetuning for adaptive vulnerability discovery.
Our framework is inspired by the concept of fuzzing (or fuzz testing) in software engineering, which expands a small set of seed inputs into many systematically varied test cases using automated templates. In a similar spirit, we propose vision-language fuzzing to generate diverse variants of a single input query in both the vision and language domains. For example, given a seed image of a ``red apple", visual perturbations may include transformations such as flipping or color adjustments, while linguistic perturbations convert the query “What color is the apple?” into variants like “What color is the apple not?” or “Is the apple red or green?”. By generating such systematic perturbations, vision-language fuzzing constructs a large and diverse family of test cases, enabling the discovery of latent vulnerabilities that would remain hidden under isolated inputs.
However, since VLM vulnerabilities are highly heterogeneous and the query space is vast, relying solely on fixed transformation templates provides limited coverage. To address this, we introduce a fuzzing model trained via adversarial reinforcement learning to adaptively explore the most failure-prone regions of the target model. During training, incorrect predictions are assigned higher rewards, driving the fuzzing model to generate increasingly challenging queries that effectively probe the weaknesses of the VLM. Through iterative refinement, FuzzingRL progressively sharpens its probing capability, leading to a targeted and incisive discovery of model failures.

\begin{figure*}[t]
  \centering
  \includegraphics[width=0.95\textwidth]{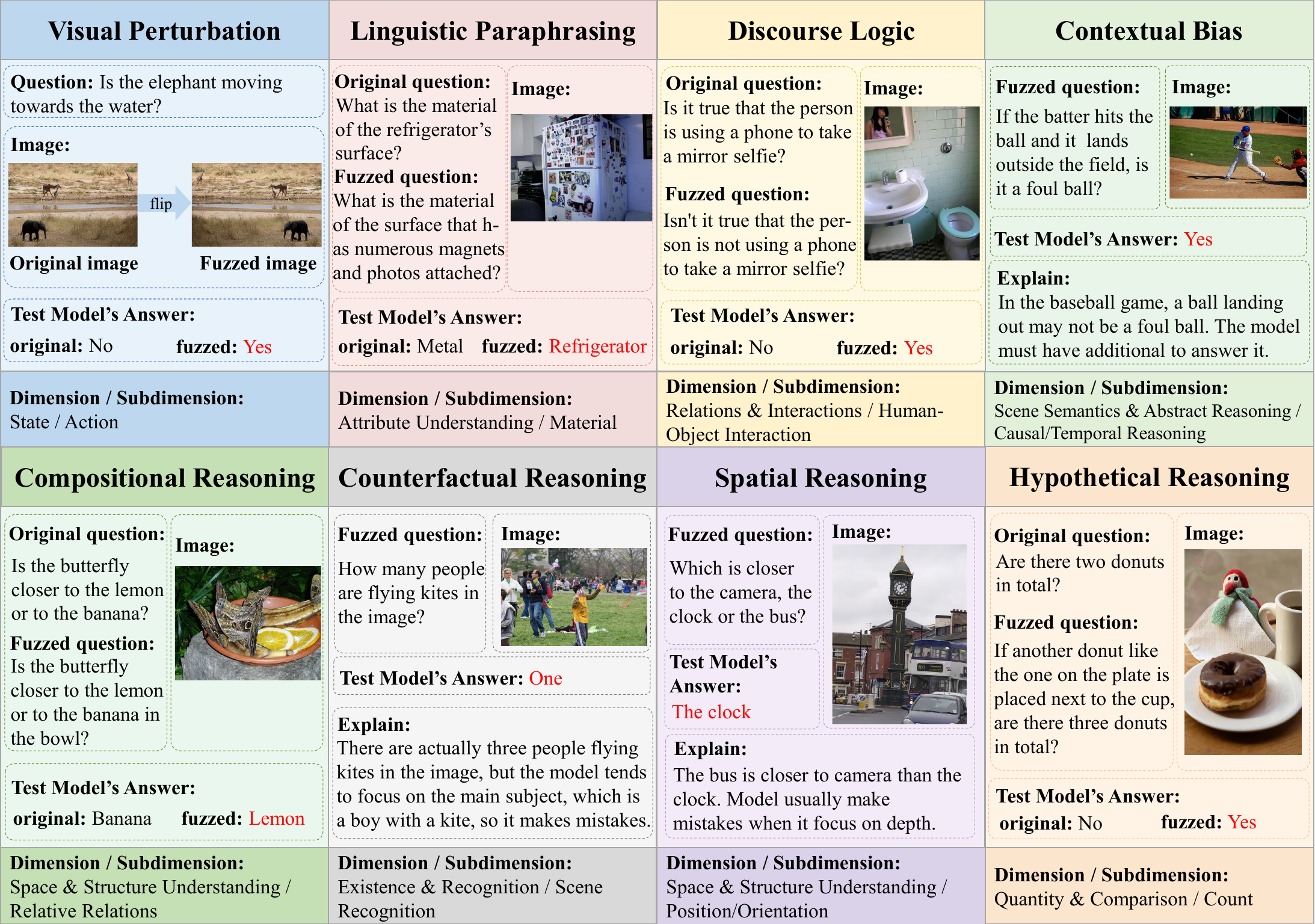}
  \vspace{-5pt}
  \caption{\textbf{FuzzingRL probe examples.} For a given capability subdimension $d$, each panel shows an answerable probe generated by a specific fuzzing role and the answer from the target model (\textit{Qwen2.5-VL-32B}).}
  \label{fig:fuzzing-examples}
  \vspace{-10pt}
\end{figure*}

The framework outputs reproducible, auditable failure cases with full metadata and aggregates them into an attributable error profile, thereby achieving automatic, reproducible, and scalable discovery and localization of systematic VLM failures with minimal human involvement under fixed budgets. We selected the Qwen2.5-VL-7B model as the fuzzing model and the Qwen2.5-VL-32B model as the target model. After four iterations of training, the target model's accuracy drops from 86.58\% to 65.53\%.

We further study transferability: after training on a single target VLM, the resulting fuzzing model can be reused to test other VLMs. Concretely, we fix the trained fuzzing model and use it to produce a held out set of image grounded probes, then query multiple test VLMs with the same questions and images and report their accuracies under a unified human annotation rubric. Across diverse architectures and scales, the generated probes consistently reduce test model accuracy. We also track this cross model effect through training iterations on a fixed panel of held-out test VLMs, and use it to select a checkpoint that avoids overfitting to the target model. Additionally, our probes reveal recurring failure patterns, including spatial reasoning, counting, and sensitivity to instruction phrasing.

\section{Related Work}

\textbf{Static Benchmarks.} Most existing VLM evaluations rely on static test suites that span general capabilities and multiple task domains: general multimodal suites \cite{liu2024mmbench,yue2024mmmu,zhang2024mme,lu2023mathvista} assess overall ability; classic vision datasets \cite{lin2014coco,gupta2019lvisdatasetlargevocabulary,Kuznetsova_2020_openimage,imagenet,Places} ground detection and scence recognition; QA datasets \cite{balanced_vqa_v2,hudson2019gqanewdatasetrealworld} anchor answers in image evidence; composition and relations \cite{suhr2019nlvr2visualbiasanalysis,yu2016modelingcontextreferringexpressions_refcoco} probe structured reasoning and alignment; counting and comparison \cite{acharya2018tallyqaansweringcomplexcounting,welde2025counting} capture number sensitivity; interaction and events \cite{chao2018hico-det,vcoco,ji2019Genome} evaluate actions and roles; test and chart understanding \cite{singh2019textvqa,biten2019stvqa,sidorov2020textcaps,mathew2021docvqa,masry2022chartqa} test reading and layout. Robustness lines of work typically apply functional perturbations (flip/rotation, crop/occlusion, compression/noise) and linguistic edits (paraphrase, negation, conditional augmentation) to test answer stability \cite{ishmam2024visualrobustnessbenchmarkvisual,zhang2025negvqa,alhamoud2025negbench,nguyen2024nopenovelobjectpose,mashrur2024robust}. However, such static evaluations share two limitations: (i) costly labeling, one-shot test banks, and different designs that weaken out of distribution comparability; (ii) little use of observed failure signals to generate new test items, making it hard to continuously expose model weaknesses. In contrast, we actively explore VLM failure-prone regions, and our test set updates dynamically as new weaknesses are discovered.

\textbf{Generative Benchmark.} To break the static bottleneck, recent work pursues programmatic dynamic test generation: Task Me Anything \cite{zhang2025taskmeanything} expands the task space via templates and asset libraries; ProVision \cite{zhang2024provision} scales up visual task; Open-ended VQA \cite{ging2024openendedvqa} automatically convert classification taxonomies into open-ended VQA; earlier CLEVR \cite{johnson2017clevr} provides a controllable synthetic pipeline. However, common gaps remains: (i) heavily relying on templates and synthetic assets, which limits transfer to real images; (ii) most method rely on a single signal, so they miss many skills from low-level perception to high-level relational reasoning; (iii) lets test composition dominate outcomes, so scores reflect question allocation rather than true model ability. In contrast, our fuzz-testing approach searches a much broader space for failures and is model-targeted, adapting tests to the specific model’s errors. Unlike coverage-oriented programmatic generators and classical coverage guided fuzzing that rely on predefined heuristics, FuzzingRL closes the loop by using failure feedback to learn a transferable fuzzing model that progressively produces harder queries across models.

\textbf{Fuzz Testing.}
Classical fuzz testing originated in systems reliability, where randomized inputs were injected into UNIX utilities to expose crashes and anomalous behavior \cite{miller1990empirical,miller1995fuzz}. Coverage-guided greybox fuzzing later scaled this paradigm by instrumenting targets and using branch-coverage feedback to steer input mutation and exploration (e.g., AFL and libFuzzer), markedly improving efficiency under tight budgets \cite{serebryany2016continuousfuzzing,bohme2017directed}. This establishs a general ``mutate--measure--explore'' loop for efficiently searching vast input spaces.

\textbf{Large Model Safety.}
Recent work adapts fuzzing style, failure seeking evaluation to large models. On the text side, adversarial suffixes and automated optimization repeatedly elicit unsafe or policy violating outputs from aligned LLMs, revealing prompt fragility and transferability \cite{zou2023adversarialattacks,wei2023jailbroken,ganguli2022red}. On the multimodal side, studies demonstrate that images can carry hidden instructions artifacts that trigger unsafe behaviors, motivating systematic safety testbeds and attack taxonomies \cite{beutel2024diverse,liu2024mmsafetybenchbenchmarksafetyevaluation,lu2024wildvision}. Together, these results underscore the limits of one-shot static suites and the need for continually generated, judgeable tests with explicit accounting of what capabilities are probed.

\section{Method}

\subsection{Overview}

Our method aims to actively find VLM failures across multiple dimensions, rather than static benchmark targeting one specific area.
At a high level, our goal is to learn the question generator $\pi_\theta$ to maximize the expected failure rate of the target VLM $f$ under diverse visual and linguistic conditions.
Formally, for images $x \in \mathcal{X}$, we define the global objective as
\begin{equation}
\label{eq:global}
\begin{aligned}
\max_{\theta}\quad
& \mathbb{E}_{x \sim \mathcal{X},\, q \sim \pi_\theta(\cdot \mid x,d,r)}
   \Big[ \underbrace{\mathcal{J}\big(x, q, f(x,q)\big)}_{\text{proxy for failure rate}} \Big] \\
\end{aligned}
\end{equation}
Here, $\mathcal{J}(x,q,\hat a) \in \{-1,0,1\}$ is the judgment signal assigning higher values to incorrect answers of target model. For the question generator $\pi_\theta(\cdot \mid x,d,r)$, $d \sim p(d)$ and $r \sim p(r)$ denote the sampling priors over fuzzing subdimensions and roles in vision-language fuzzing (Section \ref{section3-1}). This objective encourages $\pi_\theta$ to generate diverse, valid questions that maximize the likelihood of VLM failure, while remaining controllable across multiple fuzzing dimensions.
To optimize Eq.~\ref{eq:global}, we instantiate judgeability and coverage via a \emph{vision-language fuzzing} module, and then train the policy $\pi_\theta$ with \emph{adversarial reinforcement finetuning} to prefer higher-utility questions. We will introduce the vision-language fuzzing module in Section \ref{section3-1} and the adversarial reinforcement finetuning of $\pi_\theta$ in Section \ref{section3-2}.

\subsection{Vision-Language Fuzzing} \label{section3-1}

\quad Fuzz testing, or fuzzing, is a method for discovering system vulnerabilities. It works by automatically generating a large number of diverse cases to test a system, expecting to trigger and catch the failures that can indicate the weaknesses of the system. This method expands the probing space and find the failure cases automatically. In this paper, we propose a novel approach to adapt the concept of fuzzing to the realm of vision and language, termed vision–language fuzzing. Our goal is to systematically generate semantically valid and diverse variants of input queries that can reveal inconsistencies or brittleness in vision–language models.

To apply fuzz testing for VLMs, we ground our vision-language fuzzing in 24 subdimensions $d$, which map to the key abilities VLMs are commonly tested on. Across these subdimensions, we apply 8 fuzzing roles $r$ to form simple and structured variations in how questions are formed. The concrete examples that the vision-language fuzzing system generates are shown in \autoref{fig:fuzzing-examples}.

We organize 24 subdimensions into 7 capability groups (Table~\ref{tab:subdims}).
Difficulty styles are instantiated by eight fuzzing roles (Table~\ref{tab:fuzzroles});
each probe is labeled by $(d,r)$ for controllable difficulty and attribution. We introduce the fuzzing roles as follows:

\begin{table}[t]
  \footnotesize
  \centering
  \setlength{\tabcolsep}{5pt}
  \caption{Capability dimensions $d$ for stress-testing VLMs}
  \vspace{-5pt}
  \label{tab:subdims}
\setlength{\tabcolsep}{6pt}
\renewcommand{\arraystretch}{1.08}
\setlist[itemize]{leftmargin=*,label=\textbullet,topsep=1pt,itemsep=1pt,parsep=0pt}

\begin{tabularx}{\linewidth}{@{} >{\raggedright\arraybackslash}p{0.34\linewidth} >{\raggedright\arraybackslash}X @{}}
\toprule
\textbf{Capability group} & \textbf{Subdimensions} \\
\midrule

Existence \& Recognition &
\begin{minipage}[t]{\linewidth}\raggedright
\begin{itemize}
  \item Object Presence
  \item Scene Recognition
  \item Person/Animal Presence
\end{itemize}
\end{minipage}
\\[2pt]

Attributes &
\begin{minipage}[t]{\linewidth}\raggedright
\begin{itemize}
  \item Color
  \item Material
  \item Size (Relative)
  \item State/Action
\end{itemize}
\end{minipage}
\\[2pt]

Spatial \& Structural &
\begin{minipage}[t]{\linewidth}\raggedright
\begin{itemize}
  \item Position/Orientation
  \item Relative Relations
  \item Part--Whole Hierarchy
  \item Occlusion/Perspective
\end{itemize}
\end{minipage}
\\[2pt]

Quantity \& Comparison &
\begin{minipage}[t]{\linewidth}\raggedright
\begin{itemize}
  \item Count (1,2,3\ldots)
  \item Fuzzy Quantity (few/many/some)
  \item Comparison/Ranking
\end{itemize}
\end{minipage}
\\[2pt]

Relations \& Interactions &
\begin{minipage}[t]{\linewidth}\raggedright
\begin{itemize}
  \item Human--Object Interaction
  \item Human--Human Interaction
  \item Event Recognition
\end{itemize}
\end{minipage}
\\[2pt]

Scene-semantic Composition \& Abstract Reasoning &
\begin{minipage}[t]{\linewidth}\raggedright
\begin{itemize}
  \item Multi-object Reasoning
  \item Task/Scene Identification
  \item Causal/Temporal Reasoning
  \item Intention/Goal Recognition
\end{itemize}
\end{minipage}
\\[2pt]

Symbols \& Pragmatics &
\begin{minipage}[t]{\linewidth}\raggedright
\begin{itemize}
  \item Text (OCR) Recognition \& Understanding
  \item Symbol/Sign Meaning
  \item Pragmatic/Social Cues
\end{itemize}
\end{minipage}
\\
\bottomrule
\end{tabularx}

  \vspace{-8pt}
\end{table}

\begin{table}[t]
  \footnotesize
  \centering
  \setlength{\tabcolsep}{5pt}
  \caption{Fuzzing roles $r$ for generating VLM queries}
  \vspace{-5pt}
  \label{tab:fuzzroles}
\setlength{\tabcolsep}{5pt}
\renewcommand{\arraystretch}{1.08}

\begin{tabularx}{\linewidth}{@{} l >{\raggedright\arraybackslash}X @{}}
\toprule
\textbf{Role ($r$)} & \textbf{What it stresses} \\
\midrule
\texttt{Visual Perturbation}        & Robustness to light visual transforms; avoid superficial shortcuts. \\
\texttt{Linguistic Paraphrasing}    & Linguistic invariance without changing semantics (paraphrase/reorder robustness). \\
\texttt{Discourse Logic}      & Logical consistency under discourse cues (negation/entailment; polarity robustness). \\
\texttt{Contextual Bias}           & Separate visual grounding from world knowledge; resist prior-led guessing. \\
\texttt{Compositional Reasoning}& Compositionality across multiple attributes/relations within one query. \\
\texttt{Counterfactual Reasoning}    & Overcoming strong priors with visual evidence; rare or prior-violating patterns. \\
\texttt{Spatial Reasoning}     & 3D/relative depth ordering, occlusion and perspective cues. \\
\texttt{Hypothetical Reasoning}     & One–two-step numeric/logic reasoning over visible entities. \\
\bottomrule
\end{tabularx}

  \vspace{-8pt}
\end{table}

\textbf{Visual Perturbation.}
We apply semantics-preserving visual transforms (e.g., flips, mild noise) while keeping the question unchanged.
A failure is an answer change that should be invariant to such transforms (e.g., a count differs after a horizontal flip).

\textbf{Linguistic Paraphrasing.}
We rewrite the question via synonym substitution and syntactic alternations while keeping the visual evidence and answer type fixed.
A failure is sensitivity to surface form across meaning-equivalent paraphrases.

\textbf{Discourse Logic.}
We wrap a base query with discourse operators (negation, entailment, concessives) designed to preserve the intended truth conditions.
Failures include polarity flips or logically inconsistent answers under equivalent rewrites.

\textbf{Contextual Bias.}
We add plausible but unsupported commonsense distractors to test whether answers remain grounded in the image rather than priors.
A failure is an overconfident fabricated claim when the image does not provide sufficient evidence.

\textbf{Compositional Reasoning.}
We form multi-constraint queries that jointly require multiple grounded attributes and/or relations (e.g., color + position + size).
A failure is partial grounding where the model satisfies one constraint but ignores or misbinds another.

\textbf{Counterfactual Reasoning.}
We probe prior-defying yet visually explicit configurations to test whether the model overrides strong priors with direct perception.
A failure is defaulting to commonsense against clear visual evidence (e.g., insisting on ``five'' for a visible six-finger hand).

\textbf{Spatial Reasoning.}
We ask depth/occlusion/perspective questions that require 3D reasoning beyond 2D layout cues (front--back ordering, relative depth).
Failures include systematic depth confusions or treating apparent size as true distance under perspective distortion.

\textbf{Hypothetical Reasoning.}
We add lightweight conditional modifications (add/remove/recolor) that require grounded mental simulation and simple inference.
A failure is treating the condition as irrelevant or producing inconsistent grounded updates (e.g., incorrect remaining count after removal).

Equipping the question generator $\pi$ with our 8 fuzzing roles $r$ measurably increases the diagnostic pressure on the test VLM, even before the fine-tuning stage.
For instance, applying \textit{Qwen2.5-VL-72B} as the in-context question generator and \textit{Qwen2.5-VL-32B} as the test VLM, the overall answer accuracy drops from \textit{93.29\%} to \textit{85.33\%} before RFT.

\subsection{Adversarial Reinforcement Finetuning} \label{section3-2}

\begin{figure*}[t]
  \centering
  \includegraphics[width=0.98\textwidth]{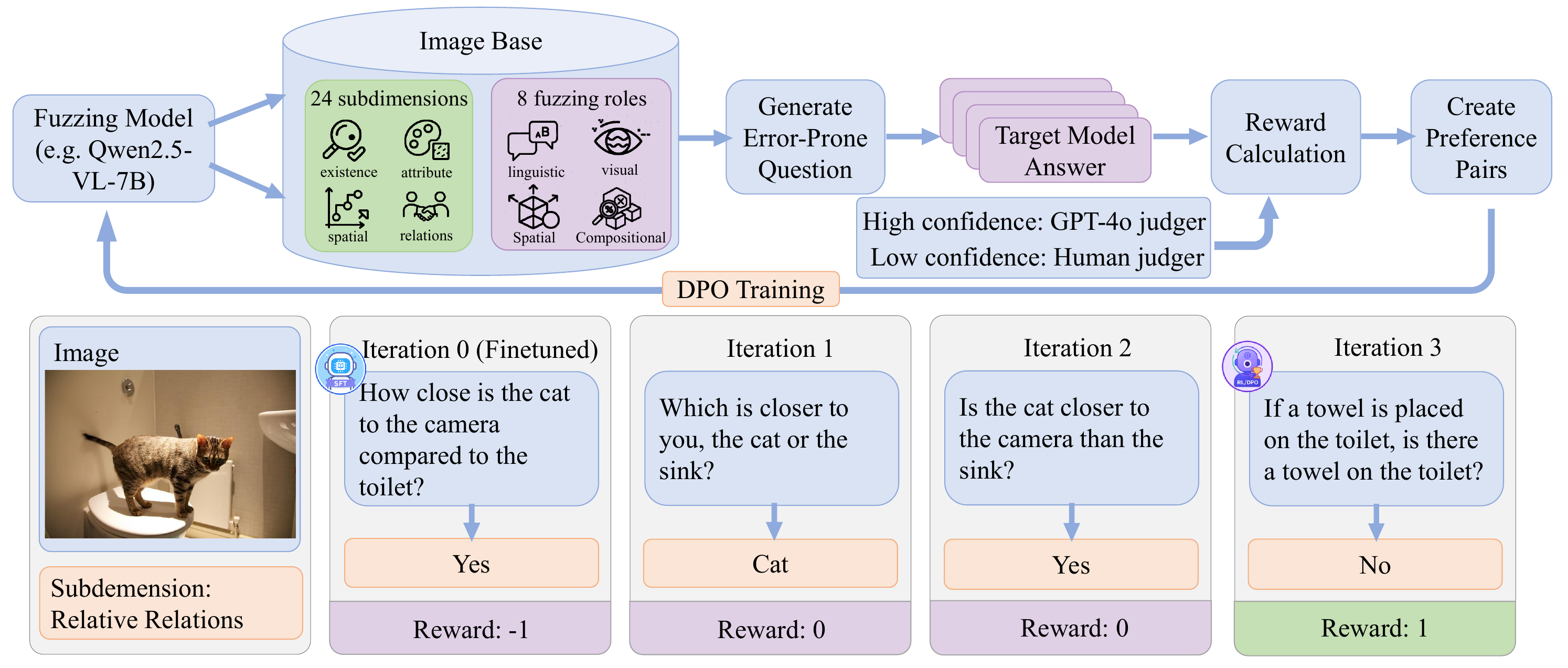}
  \vspace{-5pt}
  \caption{\textbf{Overview of FuzzingRL.} A fuzzing model (e.g., Qwen2.5-VL-7B) samples from an image base organized by 24 subdimensions and 8 fuzzing roles to generate diverse, error-prone questions for a target VLM. The target’s responses are scored via reward calculation, using a GPT-4o judge when confidence is high and a human judge otherwise, and the resulting preference pairs are used for DPO training to update the fuzzing model. The bottom panel illustrates how iterative training progressively sharpens the generated queries from ordinary perception questions to more failure-inducing, compositional prompts, thereby improving the fuzzing model’s ability to surface failure cases and making it more likely to elicit incorrect answers from VLMs.}
  \vspace{-10pt}
  \label{fig:fuzzing-training}
\end{figure*}

While the structured vision-language fuzzing system ensures broad coverage of VLM capabilities, it lacks dynamic adaptability to focus on high failure regions. Given an input image \(x\), a specified subdimension \(d\) and role \(r\), the static sampling of \(d, r\) combinations leads to inefficient resource allocation, as most queries may lead to low failure regions. To address this, we propose a novel adversarial reinforcement finetuning paradigm to train a fuzzing model that converges the query space to high failure regions. The core idea is to assign a high reward to the questions if the target VLM produces incorrect answers, thereby guiding the question generator to ask increasingly difficult questions. The whole training process is shown in \autoref{fig:fuzzing-training}.

Let \(x \in \mathcal{X}\) be an image, \(d \in \{1, \dots, 24\}\) a subdimension, and \(r \in \{1, \dots, 8\}\) a fuzzing role. The generator policy \(\pi_\theta(q \mid x,d,r)\) produces a question \(q\). The target model outputs \(\hat a = f(x,q)\). A judge \(\mathcal{J}\) assigns a ternary label \(y \in \{-1,0,1\}\) to \(\hat a\), where \(-1\) denotes \emph{unanswerable}, \(0\) denotes \emph{correct}, and \(1\) denotes \emph{incorrect}. In our setup, \(\mathcal{J}\) is a committee composed of GPT-4o and human judges: we query GPT-4o five times and collect its self reported confidence; if the majority label has agreement $\ge 80\%$ and all majority runs have confidence $\ge 0.90$, we accept the GPT-4o majority label as \(y\), otherwise we defer to a human judge. We validate this gating by measuring agreement with human annotations, achieving 88.12\% agreement for the high-confidence subset and 61.14\% for the low-confidence subset. To more robustly train the question generator \(\pi_\theta\), we propose a novel three-step learning paradigm:

\textbf{SFT bootstrapping.} We supervise \(\pi_\theta\) on two complementary synthetic batches to obtain a format-ready, role-controllable initializer \(\pi_0\) (frozen as \(\pi_{\mathrm{ref}}\)).
Batch A (\textit{coverage}) exhaustively instantiates all \(24\times 8\) \((d,r)\) on a seed image set to learn strict formats and role control;
Batch B (\textit{preference hints}) picks one plausible role per \(d\) on a larger image set to provide weak role-selection cues without overfitting.
The SFT loss is
\begin{equation}
\label{eq:sft}
\mathcal{L}_{\mathrm{SFT}}
= - \mathbb{E}_{(x,d,r,q^\star)} \big[\log \pi_\theta(q^\star \mid x,d,r)\big]
\end{equation}
We set \(\pi_0=\arg\min_\theta \mathcal{L}_{\mathrm{SFT}}\) and \(\pi_{\mathrm{ref}}=\pi_0\).

\textbf{In-context preference construction.} To neutralize content difficulty, we fix the context \((x,d)\) and vary the fuzzing role \(r\) and wording.
Here \(r\in\{1,\dots,8\}\) denotes the fuzzing role. We sample \(N\) candidates
\begin{equation}
q_i \sim \pi_\theta(\cdot \mid x,d,r_i), \quad r_i \in \{1,\dots,8\}, \;\; i=1,\dots,N,
\end{equation}
and query the target VLM for answers
\begin{equation}
\hat a_i = f(x,q_i).
\end{equation}
Then, we let the committee judges assign ternary labels
\[
\begin{aligned}
y_i &\in \{-1,0,1\} \\
&\text{(\(-1\)=unanswerable,\; \(0\)=correct,\; \(1\)=incorrect).}
\end{aligned}
\]
We score each candidate by
\begin{equation}
\label{eq:score-basic}
s_i \;=\; y_i \;\in\; \{-1,0,1\},
\end{equation}
and form a in-context preference pair simply as
\begin{equation}
\label{eq:pair}
q^{+} \in \arg\max_i s_i, \qquad
q^{-} \in \arg\min_i s_i ,
\end{equation}
which constructs preferences that primarily reflect role selection and phrasing under the same \((x,d)\).

\textbf{Direct preference optimization.} Fixing \(\pi_{\mathrm{ref}}\), we optimize \(\pi_\theta\) with direct preference optimization (DPO) such that under the same \((x,d)\), $\pi_{\theta}$ prefers questions with higher failure scores (including unanswerable).
Let \((q^{+},q^{-})\) be the contrastive pairs in Eq.~\ref{eq:pair}.
Our DPO objective is
\begin{equation}
\label{eq:dpo}
\begin{aligned}
\mathcal{L}_{\mathrm{DPO}}
&= - \mathbb{E}_{(x,d,q^{+},q^{-})}
   \Big[\log \sigma\!\big(\beta\,\Delta_{\theta,\mathrm{ref}}(x,d,q^{+},q^{-})\big)\Big] \\
&\quad + \lambda_{\mathrm{KL}}\,\mathbb{E}_{(x,d)}
   \Big[\mathrm{KL}\!\big(\pi_\theta(\cdot|x,d)\,\|\,\pi_{\mathrm{ref}}(\cdot|x,d)\big)\Big],
\end{aligned}
\end{equation}
where $\Delta_{\theta,\mathrm{ref}}$ is
\begin{equation}
\begin{aligned}
\Delta_{\theta,\mathrm{ref}}
&= \big(\log \pi_\theta(q^{+}|x,d) - \log \pi_\theta(q^{-}|x,d)\big) \\
&\quad - \big(\log \pi_{\mathrm{ref}}(q^{+}|x,d) - \log \pi_{\mathrm{ref}}(q^{-}|x,d)\big) .
\end{aligned}
\end{equation}

We repeat the preference construction and optimization stages multiple times. After several iterations, the updates stabilize and yield a trained fuzzing model that concentrates generation on the target VLM's most failure-prone regions.

\section{Experiments}
\label{sec:experiments}

In this section, we investigate the efficacy of FuzzingRL for automatically detecting VLM failures, and analyze key design choices and limitations. To this end, we organize our study to answer the following empirical questions, in order:

\textbf{(Q1) Overall Performance:} How effective and efficient is FuzzingRL in revealing VLM failure modes?

\textbf{(Q2) Vision-Language Fuzzing:} To what extent does our vision–language fuzzing strategy broaden the test space and improve the discovery of diverse failures?

\textbf{(Q3) Adversarial Reinforcement Fine-Tuning:} Can adversarial RFT successfully steer the question generator toward increasingly challenging and failure-inducing queries?

\textbf{(Q4) Data Analysis:} What insights can we obtain from analyzing the collected failure cases, and what do these imply for future VLM robustness research?

\subsection{Implementation Details}
For image queries to VLMs, we sample them from the COCO dataset~\cite{lin2014coco}.
Evaluation follows our 24-subdimension taxonomy. Unless noted, each method produces one \emph{annotators-pass} probe per (image $x$, subdimension $d$).

\textbf{Evaluating a question generator.}
We evaluate a question generator by letting it sample images from the same validation image pool and generate one probe per (image $x$, subdimension $d$) under identical decoding conditions (temperature, top-$p$, and max tokens).
We then query a fixed \emph{test model} (GPT-4o) with each (image, question) pair, and human judges annotate the test model's outputs for correctness and answerability under a unified rubric.

\textbf{Metrics.}
We report three diagnosis-oriented metrics on \emph{annotators-pass} items:

Fooling rate $FR$ ($=1-\mathrm{Acc}$ of the test model), where a higher value indicates that the question generator more effectively hits the test model’s weak spots.

Unanswerable rate $UR$: fraction of generated probes that are unanswerable from the image (as judged by the human annotators); lower $UR$ indicates cleaner, image-grounded probing.

Distinct Ratio $DR$: measuring per-image uniqueness and diversity of generated questions.
For each image $i$, let $Q_i$ be the set of generated questions and $\operatorname{uniq}(Q_i)$ the de-duplicated set. We define
\[
\mathrm{DR}=\frac{1}{|I|}\sum_{i\in I}\frac{\bigl|\operatorname{uniq}(Q_i)\bigr|}{\lvert Q_i\rvert},
\]
where $I$ is the set of images. A higher $\mathrm{DR}$ indicates fewer templated duplicates and broader variation.

\subsection{Overall Performance of FuzzingRL}
Following the above evaluation protocol, we compare our trained FuzzingRL generator against Qwen2.5-VL, Llama-3.2-Vision and GPT-4o as question generators, each without an explicit fuzzing role prompt.
In all cases, we keep the evaluation test model fixed (GPT-4o) and compute $FR/UR/DR$ from human annotations.
\Cref{tab:main} summarizes results.

\begin{table}[t]
  \centering
  \small
  \setlength{\tabcolsep}{10pt}
  \caption{\textbf{Overall performance of FuzzingRL.} Applying FuzzingRL to the small generator Qwen2.5-VL-7B enhances fooling rate from 0\% to 34.47\% while keeping the unanswerable rate low and diversity competitive. Remarkably, a small base model Qwen2.5-VL-7B + FuzzingRL outperforms the Llama-3.2-11B, the much larger models Qwen2.5-VL-72B and the closed-source GPT-4o, effectively turning a weak base generator into a strong vulnerability finder.}
  \vspace{-5pt}
  \label{tab:main}
  \centering
\resizebox{0.9\linewidth}{!}{%
\begin{tabular}{lccccc}
    \toprule
    Question Generator & FR $\uparrow$ & UR $\downarrow$ & DR $\uparrow$ \\
    \midrule
    Qwen2.5-VL-7B             & 0.00\%  & 100.00\% & -- \\
    Qwen2.5-VL-72B            & 6.71\%  & 9.33\% & 95.83\% \\
    Llama-3.2-11B-Vision      & 1.00\%  & 12.84\% & 93.98\% \\
    GPT-4o                   & 7.59\%  & 4.50\%   & 91.37\% \\
    \textbf{Qwen2.5-VL-7B + fuzzingRL} & \textbf{34.47}\% & \textbf{7.75}\% & \textbf{91.50}\% \\
    \bottomrule
\end{tabular}
}
  \vspace{-10pt}
\end{table}

\subsection{Vision-Language Fuzzing}
We ablate the contribution of the vision-language fuzzing design by comparing generation \emph{with} versus \emph{without} fuzzing roles used as few-shot in-context exemplars, while keeping the test model and decoding budget fixed. \Cref{tab:VLfuzzing} reports both the absolute numbers and the deltas \((\Delta=\text{with}-\text{without})\).

\begin{table}[t]
  \centering
  \small
  \setlength{\tabcolsep}{10pt}
  \caption{\textbf{Effect of Vision-Language Fuzzing.} We evaluate whether vision-language fuzzing itself, used as few-shot in-context exemplars without RFT, can improve the quality of generated test queries.
  We report deltas between \emph{with} and \emph{without} fuzzing (\(\Delta = \text{With fuzzing} - \text{Without fuzzing}\)) for $FR$, $UR$, and $DR$.
  Across different models, vision-language fuzzing consistently increases the fooling rate and the diversity of generated questions while reducing unrealistic queries.
  For a fair comparison with the \emph{without fuzzing} setting, we keep the generation budget identical and generate exactly one question per image in the \emph{with fuzzing} setting as well.}
  \vspace{-5pt}
  \label{tab:VLfuzzing}
  
\centering
\resizebox{0.9\linewidth}{!}{%
\begin{tabular}{@{}llccc@{}}
    \toprule
    Model & Setting & FR $\uparrow$ & UR $\downarrow$ & DR $\uparrow$ \\
    \midrule
    Qwen2.5-VL-72B & w/o fuzzing &  6.71\% &  9.33\% & 95.83\% \\
    Qwen2.5-VL-72B & w/ fuzzing  & 14.67\% &  2.33\% & 98.75\% \\
    \textbf{$\Delta$} & & \textbf{+7.96} & \textbf{-7.00} & \textbf{+2.92} \\
    \midrule\midrule
    Llama-3.2-11B-Vision & w/o fuzzing & 1.00\% & 12.84\% & 93.98\% \\
    Llama-3.2-11B-Vision & w/ fuzzing  & 3.61\% & 5.23\% & 94.39\% \\
    \textbf{$\Delta$} & & \textbf{+2.61} & \textbf{-7.61} &
    \textbf{+0.41} \\
    \midrule\midrule
    GPT-4o & w/o fuzzing & 7.59\% & 4.50\% & 91.37\% \\
    GPT-4o & w/ fuzzing  & 9.62\% & 4.75\% & 93.08\% \\
    \textbf{$\Delta$} & & \textbf{+2.03} & \textbf{+0.25} & \textbf{+1.71} \\
    \midrule\midrule
    FuzzingRL (Ours) & w/o fuzzing & 31.26\% & 7.67\% & 91.13\% \\
    FuzzingRL (Ours) & w/ fuzzing  & 34.47\% & 7.75\% & 91.50\% \\
    \textbf{$\Delta$} & & \textbf{+3.21} & \textbf{+0.08} & \textbf{+0.37} \\
    \bottomrule
\end{tabular}%
}


  \vspace{-15pt}
\end{table}

Across all generators in \Cref{tab:VLfuzzing}, adding vision-language fuzzing roles to the few-shot context consistently increases FR, indicating that role prompts impose a structured prior over question space and steer generation toward systematic, error-revealing probes rather than benign queries. This shift is not driven by more unanswerable outputs: UR stays stable or decreases, suggesting that roles constrain generation toward answerable questions that test models attempt rather than reject. DR also increases with roles, because roles provide distinct templates that diversify how the same subdimension is probed and reduce near-duplicate questions per image. Overall, roles act as a structured exploration prior, expanding surface realizations while keeping questions answerable and on-topic.

Combining vision-language fuzzing with reinforcement learning (FuzzingRL) produces a clear synergy. Vision-language fuzzing diversifies the probing space in a controlled, answerable manner, and RL further concentrates probability mass on hard yet valid queries. This combination preserves the higher FR and typically increases DR by reducing template reuse and encouraging distinct, role-consistent realizations. Overall, FuzzingRL strengthens failure discovery while keeping UR low, improving both diagnostic power and diversity at the same time.

\begin{figure}[t]
  \centering
  \includegraphics[width=0.90\linewidth]{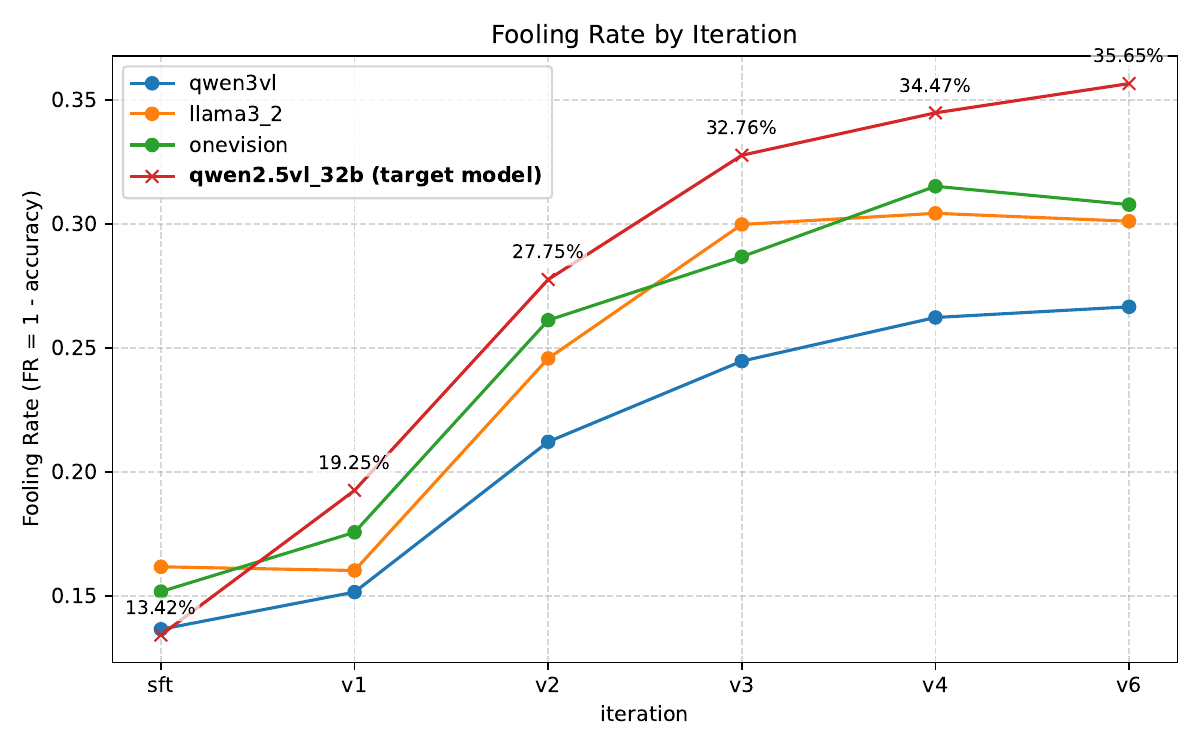}
  \vspace{-5pt}
  \caption{\textbf{FR by iteration with transfer evaluation.}
  FR ($=1-\mathrm{Acc}$) over training iterations on the target model (Qwen2.5-VL-32B) and three held-out test VLMs.
  FR on the target increases steadily, whereas transfer FR peaks around iteration~4 and may drop with further training, so we stop at iteration~4.}
  \vspace{-20pt}
  \label{fig:iter}
\end{figure}

\begin{figure*}[t]
  \centering
  \includegraphics[width=0.95\linewidth]{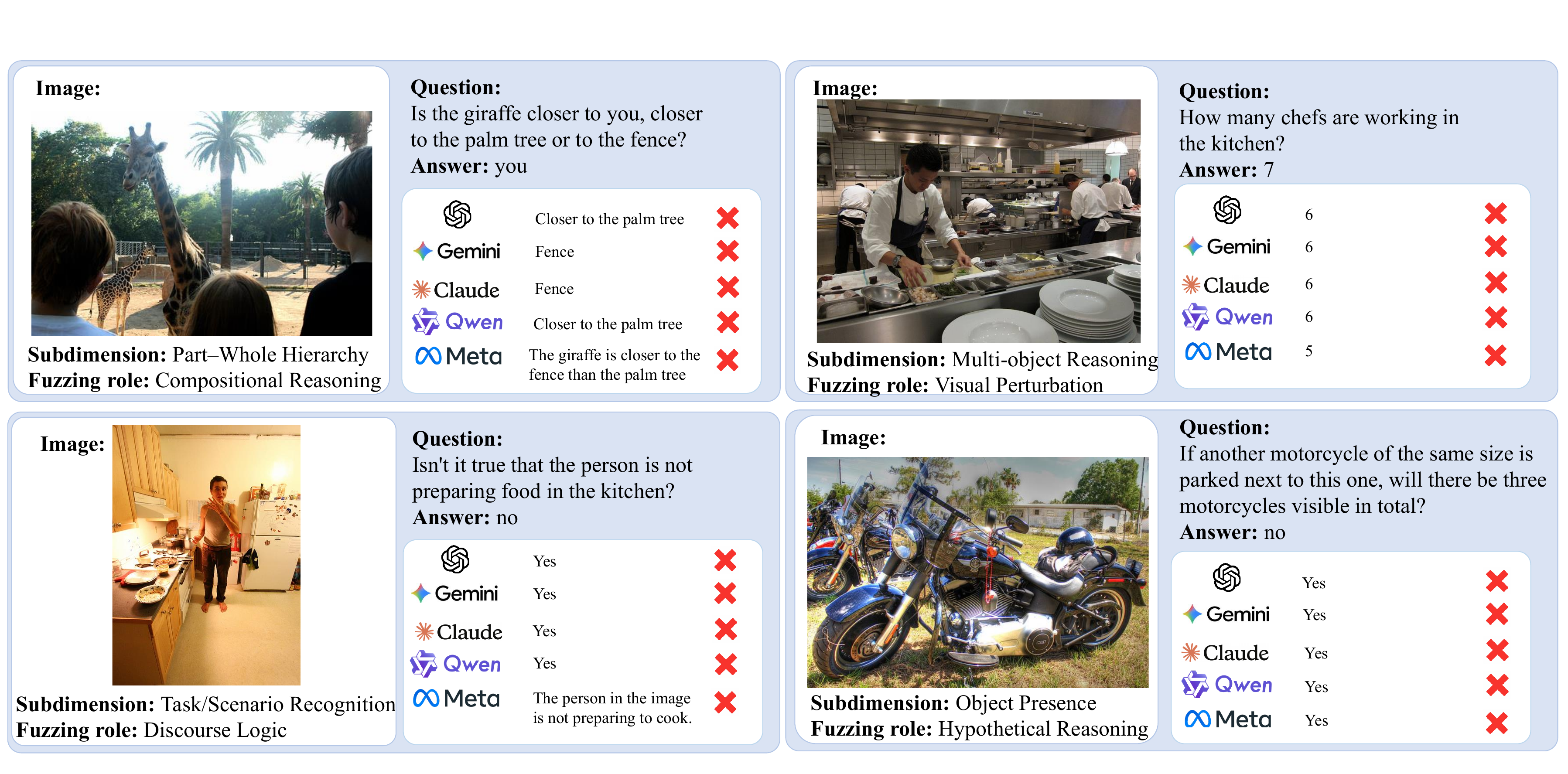}
  \vspace{-5pt}
  \caption{\textbf{Cross-model results of the trained fuzzing model.}
  Example probes and test-model outputs showing four recurrent failure families: (a) part--whole/compositional proximity, (b) counting under clutter, (c) discourse/negation in scenario recognition, and (d) hypothetical object-presence reasoning.}
  \vspace{-15pt}
  \label{fig:fb_example}
\end{figure*}

\subsection{Adversarial Reinforcement Fine-Tuning}
We fine-tune the fuzzing model with multiple adversarial RL rounds. In each round the policy proposes questions, the fixed target model answers, a verifier judges success and validity, and the policy is updated from the online payoff. Iteration~0 is the SFT initializer. \Cref{fig:iter} reports the fooling rate (FR $=1-\mathrm{Acc}$) of the generated probes across training iterations, evaluated on the training target model and several held-out test VLMs.

We observe that FR on the target model increases monotonically as training proceeds, indicating that the fuzzing generator becomes progressively better at eliciting failures on the target.
However, transfer performance on held-out test VLMs saturates and can even degrade after roughly four iterations, suggesting over-specialization to the target.
Therefore, we stop at iteration~4 as our final checkpoint to balance target effectiveness and cross-model generalization.

\subsection{Generalization: Stress Testing Diverse VLMs with a Trained Fuzzing Generator}
\label{sec:generalization_testmodels}

Beyond degrading a single target model during training, we study whether a \emph{trained} fuzzing generator can serve as a reusable stress-testing tool for other VLMs.
Concretely, we fix a trained fuzzing model (our final FuzzingRL checkpoint) and use it to generate one probe per (image $x$, subdimension $d$) on a held-out set of $N$ COCO images, following the same 24-subdimension taxonomy.
We then evaluate a set of \emph{test} VLMs by querying each model with the same generated questions and images, and compute their accuracy under identical evaluation rules.

To attribute the gain to FuzzingRL rather than to question formatting artifacts, we compare three generators under the same budget: (i) a base generator without adversarial fine-tuning, (ii) the same base generator augmented with our vision-language fuzzing roles as in-context exemplars (\S\ref{section3-1}) but \emph{without} RFT, and (iii) our trained FuzzingRL generator.
Importantly, questions are generated \emph{once} by each generator and are not selected or filtered based on any test model's responses, avoiding cherry-picking effects.

\begin{wraptable}{r}{0.5\linewidth}
\raggedright
\centering
  \small
  \setlength{\tabcolsep}{0pt}
  \vspace{-10pt}
  \caption{\textbf{Generalization across test VLMs.} We use the trained FuzzingRL generator to produce one probe per (image $x$, subdimension $d$) on $N$ held-out COCO images, and evaluate multiple test VLMs on the resulting question set. Answer correctness is annotated by humans. Lower accuracy indicates that the generated questions are harder for the test model.}
  \label{tab:generalization}
  \resizebox{0.80\linewidth}{!}{%
\begin{tabular}{lc}
  \toprule
  Test Model & Acc ($\uparrow$) \\
  \midrule
  Human                     & \textit{100.00\%} \\
  GPT-4o                    & \textit{83.86\%} \\
  Gemini-1.5-Flash (V)      & \textit{78.71\%} \\
  Qwen3-VL-32B              & \textit{73.78\%} \\
  Llama-3.2-Vision-11B      & \textit{69.58\%} \\
  LLaVA-OneVision-1.5-8B    & \textit{67.45\%} \\
  \bottomrule
\end{tabular}
}


\end{wraptable}

\Cref{tab:generalization} shows that questions generated by FuzzingRL consistently reduce the accuracy of diverse test VLMs compared to both baselines, indicating strong cross-model generalization of failure-seeking behaviors.
Meanwhile, the unanswerable rate remains low, suggesting that the difficulty increase is not driven by invalid or image-ungrounded questions, but by systematically probing challenging visual reasoning patterns that VLMs tend to fail on.
Overall, these results support the use of our trained fuzzing generator as a transferable stress-testing mechanism: once trained, it can be directly applied to other VLMs to elicit harder, failure-inducing queries under a fixed evaluation budget.

\subsection{Findings}
Through our trained fuzzing model, we identify several recurring failure patterns:

\textbf{Different subjects.}
Even when the intended semantics remain the same, changing the subject or reference frame of a question can lead to different answers.
For example, rewriting “Which one is closer to you?” as “Which one is closer to the camera?” may flip the prediction, despite the two being semantically equivalent in context.
This suggests sensitivity to superficial phrasing cues rather than stable grounding in the visual evidence.

\textbf{Yes/No Questions.}
For the same underlying query, rephrasing an alternative question into a yes/no form can change the model’s response.
We observe a tendency to over-predict “Yes” on some binary questions, making such formulations more likely to induce incorrect answers.
This suggests a systematic yes-bias that can outweigh visual evidence.

\textbf{Add Additions.}
The model can be disrupted by additional constraints or conditions, even when the resulting question remains simple for humans.
For instance, changing “How many cars are there here?” to “If 20 cars identical to the one on the left in the image were added, how many cars would there be here?” often yields an incorrect count.
This points to brittleness under compositional conditionals and grounded arithmetic.

\textbf{High Count Answer.}
For counting queries, performance is generally reliable when the number of target objects is small; however, once the count exceeds five, accuracy drops sharply.

\section{Conclusion}

Drawing inspiration from software fuzz-testing, we introduced FuzzingRL, a reinforcement-driven framework for the automatic discovery of modern VLM failures. Through our study, we demonstrate that VLM vulnerabilities can be progressively surfaced by refining a question generator that actively seeks out high-failure regions of the model's input space. By combining structured vision–language fuzzing with adversarial reinforcement finetuning, FuzzingRL consistently amplifies failure-inducing behaviors, driving down model accuracy over subsequent iterations and revealing weaknesses that static benchmarks cannot capture. To demonstrate our method's transferability to other VLMs, we apply the trained fuzzing generator to multiple held-out test VLMs and observe consistent accuracy drops on the same image-grounded probes, revealing systematic weaknesses in spatial reasoning, compositionality, discourse logic, and multi-object understanding.

\bibliographystyle{splncs04}
\bibliography{main}

\end{document}